\documentclass[nohyperref]{article}

\usepackage{microtype}
\usepackage{graphicx}
\usepackage{subfigure}
\usepackage{booktabs} 
\usepackage{enumitem}
\usepackage{wrapfig}
\usepackage[table,xcdraw]{xcolor}
\usepackage{hyperref}

\usepackage[accepted]{icml2022}

\usepackage{amsmath}
\usepackage{amssymb}
\usepackage{mathtools}
\usepackage{amsthm}

\usepackage[capitalize,noabbrev]{cleveref}

\theoremstyle{plain}
\newtheorem{theorem}{Theorem}[section]

\theoremstyle{definition}

\newtheorem{assumption}[theorem]{Assumption}
\newtheorem{conjecture}[theorem]{Conjecture}
\theoremstyle{remark}

\usepackage[textsize=tiny]{todonotes}

\DeclareMathOperator*{\argmax}{arg\,max}
\DeclareMathOperator*{\argmin}{arg\,min}

\icmltitlerunning{Repeated Environment Inference for Invariant Learning}

\begin{document}

\twocolumn[
\icmltitle{Repeated Environment Inference for Invariant Learning}



\icmlsetsymbol{equal}{*}

\begin{icmlauthorlist}
\icmlauthor{Aayush Mishra}{jhu}
\icmlauthor{Anqi Liu}{jhu}
\end{icmlauthorlist}

\icmlaffiliation{jhu}{Department of Computer Science, Johns Hopkins University}

\icmlcorrespondingauthor{Aayush Mishra}{amishr24@jh.edu}

\icmlkeywords{Machine Learning, Invariant Learning, ICML}

\vskip 0.3in
]



\printAffiliationsAndNotice{} 

\begin{abstract}
We study the problem of invariant learning when the environment labels are unknown. We focus on the invariant representation notion when the Bayes optimal conditional label distribution is the same across different environments. Previous work conducts Environment Inference (EI) by maximizing the penalty term from Invariant Risk Minimization (IRM) framework. The EI step uses a reference model which focuses on spurious correlations to efficiently reach a good environment partition. However, it is not clear how to find such a reference model. In this work, we propose to repeat the EI process and retrain an ERM model on the \textit{majority} environment inferred by the previous EI step. Under mild assumptions, we find that this iterative process helps learn a representation capturing the spurious correlation better than the single step. This results in better Environment Inference and better Invariant Learning.  We show that this method outperforms baselines on both synthetic and real-world datasets.
\end{abstract}
\section{Introduction}
\label{submission}



In conventional machine learning, training data is assumed to be independently and identically distributed (\textit{iid}) as the test data. This assumption is usually violated in the real world. Samples used in training might not be representative of the whole data distribution in many applications. Therefore, performance suffers when machine learning algorithms are deployed in new \textit{domains} or \textit{environments} where spurious correlations learned from the training data do not hold. To solve this problem, recent literature advocates focusing on extracting causal relationships \cite{peters2017elements, 7tools} from data. However, causal discovery and inference in high-dimensional data remain a big challenge in practice \cite{towardscrl}.

\begin{figure}[h]
	\centering
	  \includegraphics[width=0.99\linewidth]{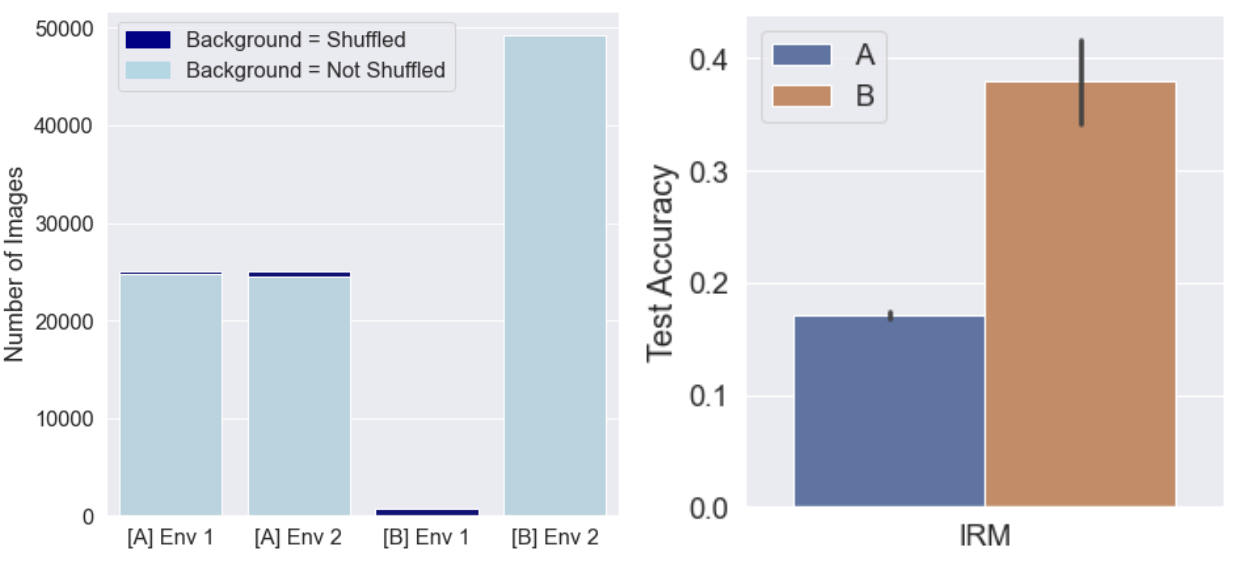}\\
	\caption{\textbf{Left:} Data distributions in two environments in experiment A and B; \textbf{Right:} IRM test accuracy in experiment A and B;
	Experiment \textbf{A}: CBMNIST [I] samples with \textit{shuffled} background w.p. 1\% and 2\% in both environments, respectively; 	Experiment \textbf{B}: Samples with \textit{shuffled} background are all grouped in one environment. The test environment has completely different backgrounds from training in both cases. IRM's performance is significantly higher in Experiment B. More details in Section \ref{sec:datasets}.}
	\label{fig:irm_diff}
\end{figure}

Recent works \cite{irm, diva, chuang} have proposed to learn stable correlations that are invariant across different domains because invariance has shown strong links with causality \cite{icp, ciip}.
However, invariant learning methods usually require predefined environment labels, which are not always available. For example, fair machine learning can be regarded as an invariant learning problem where different sub-populations correspond to different environments, and the sensitive features are usually unavailable or even hard to define \cite{corbett2018measure}. On the other hand, environment labels play an important role in invariant learning. Figure \ref{fig:irm_diff} shows how the performance of IRM \cite{irm} increases when training samples are grouped differently (Experiment B) from the ground-truth group labels (Experiment A). To create two training environments, we randomly \textit{shuffle} the background for digits with different probabilities. IRM benefits from the regrouping as the new grouping reflects a better differentiation between the two environments. See Figure \ref{fig:irm_diff}. 

\citet{eiil} proposed EIIL, enabling invariant learning when environment labels are unavailable. It conducts Environment Inference (EI) using the representation learned by a reference model by maximizing the penalty term in the Invariant Risk Minimization (IRM) framework. Moreover, it is shown that EIIL usually works when the reference model captures ``spurious" correlations. However, it is unclear how to find such a model besides learning an ERM model from all the training data.

In this paper, we find that environment partition obtained in a single EI step after ERM is often sub-optimal for invariant learning. We propose a simple repeated EI strategy to improve the quality of environment inference and, consequently, the performance of Invariant Learning. \textbf{Our contribution can be summarized as follows:}
\begin{enumerate}[leftmargin=13pt,topsep=0pt,itemsep=0mm]
\item We propose a novel method that generates environment labels by repeating the EI step and training the ERM model on the \textit{majority} environment. We call it Repeated Environment Inference for Invariant Learning (REIIL).
\item Under mild assumptions, we show that our strategy better facilitates downstream invariant learning. 
\item Our approach outperforms previous methods, including EIIL on previously used and a newly introduced dataset.
\end{enumerate}


\section{Related Work}

\textbf{IRM and Environment Inference}
\citet{irm} propose a training objective for learning the invariant representations, under the assumption that the Bayes optimal conditional $P(y|\phi(x))$ remains invariant across domains.  However, recent works \cite{rirm, kamath} have shown that IRM fails when target domains are not sufficiently similar to training domains. Besides, it also requires good environment labels beforehand. Several methods have been proposed to infer the environment labels before or along with the invariant learning. EIIL \cite{eiil} takes a two-step strategy and first partitions the training data into environments using a biased reference model before using environment-based invariant learning. However, it is unclear how to get a perfectly biased reference model to obtain a good environment partition. Adversarial Invariant Learning \cite{ail} uses a minimax game to robustly train the predictors while inferring the worst-case environment partition. In this paper, we follow the EIIL setup and study how to better conduct environment inference for downstream invariant learning.


%

\textbf{Other Types of Invariance in Domain Generalization}
Besides IRM, other definitions of \textit{invariance} exist in the domain generalization literature. Dating back to early works in domain adaptation \cite{gretton2009covariate,sun2016deep}, marginal feature alignment motivates the learning of the domain-invariant features, that is $P(\phi(x))$ being invariant across domains \cite{ganin}. However, the shortcoming of these methods is also obvious: it ignores the relationship between the features and the labels, so it still suffers under label shift \cite{zhao19,zhao2020fundamental}. 
Other works utilize tools like Information Bottleneck and propose learning objectives based on the mutual information \cite{iib}. \citet{yli} focus on conditional domain invariant representation and make sure that the $P(\phi(x)|y)$ remains invariant across domains. Moreover, focusing on only one type of invariance may be inadequate \cite{shui2022benefits}, and choosing the right type of invariant learning algorithm can be tricky.

Some works also propose using domain-specific features along with domain-invariant features to improve performance in domain generalization tasks \cite{fu, chatto, manh}. However, it is hard to say whether exploiting domain-specific features will always help in unseen domains without theoretical guarantees. Our paper aims to learn the invariant $P(y|\phi(x))$ distribution when environment labels are unavailable.


\textbf{Representation Learning under Domain Shift}
Other representation learning techniques have also been found helpful under domain shift. For example, \citet{jtt} proposes that upweighting underperforming samples in a second iteration of training is sufficient to improve worst-case group performance. \citet{lastlayer} suggest that last layer retraining of Neural Networks with interesting samples might be sufficient for robustness against spurious correlations. Contrastive Learning approaches \cite{contrast, kim2021selfreg} have also shown promise for domain adaptation and generalization. However, these methods do not explicitly learn any invariance and are usually hard to analyze. In our paper, we focus on the more principled IRM setting and overcome the difficulty of obtaining biased models for environment inference.

\section{Method}
\subsection{IRM method and Environment Inference}
The practical version of IRM (called IRMv1) has the following objective,
\begin{equation}
    \min_{\Phi:\mathcal{X}\to\mathcal{Y}}\sum_{e\in\mathcal{E}_{tr}} R^e(\Phi) + \lambda\cdot\lVert\nabla_{w\vert w=1.0}R^e(w\cdot\Phi)\rVert^2
\end{equation}
where samples are taken from multiple environments and this additional information is used to regularize training. \\
As multiple environments are not always available, EIIL proposes an environment inference step to produce the required environment labels. It uses a fixed reference classifier $\Phi$ and the IRM penalty term to do this. The \textbf{EI step} maximizes the following with respect to soft environment assignment $\boldsymbol{q}$,
\begin{equation}
    \label{eq:ei}
    C^{EI}(\Phi, \boldsymbol{q}) = \lVert\nabla_{w\vert w=1.0}\Tilde{R}^e(w\cdot\Phi, \boldsymbol{q})\rVert^2,
\end{equation}
where
    ${R}^e(\Phi, \boldsymbol{q}) = \frac{1}{N}\sum_i\boldsymbol{q}_i(e)l(\Phi(x_i), y_i)$.

This soft assignment is used to produce hard environment labels which are then used for invariant learning.\\

In summary, EIIL takes the following steps:
\begin{enumerate}[leftmargin=13pt,topsep=0pt,itemsep=0mm]
    \item Takes an input reference model $\Tilde{\Phi}$.
    \item Maximizes $C^{EI}$ to get environment partition. $\boldsymbol{q}^* = \argmax_q C^{EI}(\Tilde{\Phi}, \boldsymbol{q})$.
    \item Minimizes $C^{IL}$ (any Invariant Learning objective of choice) to get the final model. $\Phi^* = \argmin_\Phi C^{IL}(\Phi, \boldsymbol{q}^*)$.
\end{enumerate}
Note that here $\boldsymbol{q}$ is used to produce a \textbf{binary partition}, and we assume this binary partitioning throughout this paper.\\ 

EIIL works under the assumption that the reference model ($\Tilde{\Phi}$) focuses only on the spurious features. It is only when $\Tilde{\Phi} = \Phi_{Spurious}$, does the environment inference (EI) process maximally violates the Invariance Principle (EIC): 
{\small
\begin{align*}
    \mathbb{E}[y | \Phi(x) = h, e_1] = \mathbb{E}[y | \Phi(x) = h, e_2], \forall h \in \mathcal{H}, e_1, e_2 \in \mathcal{E}^{obs}
\end{align*}
}

The authors use $\Phi_{ERM}$ as an approximation for $\Phi_{Spurious}$, but this does not always work. As underlined in the paper explicitly via various experiments, it remains unexplored to find better reference models (which would be worse than ERM in terms of generalization performance) to fully exploit the EI step in EIIL. 

\subsection{Proposed method}
We present an assumption about the training distribution and a conjecture about the resulting EI step using a model trained from such a training distribution. We then reason how finding a model that satisfies the conjecture under the assumption would help the downstream invariant learning. We finally propose a strategy to approximate such a model.
\assumption 
\label{ass:ass1}
Assume a data generating graph $X \to Y \to Z$, where $X$ and $Z$ are observed features, $Y$ is the target feature and the anti-causal mechanism is unstable \cite{subbaswamy2022unifying}. A training dataset is sampled following this graph where anti-causal features are assumed to be more informative than the causal features, i.e., 
\begin{equation}
    I_{tr}(Y;Z) > I_{tr}(Y;X) > 0
\end{equation}
where $I$ measures \textit{mutual information} between features.
\conjecture
\label{con:con1}
\textit{There exists a model trained on such training datasets ($tr$) such that when it is used as a reference model for the EI step (\ref{eq:ei}), the resulting majority environment ($e_{maj}$) satisfies the following:}  
\begin{equation}
    I_{e_{maj}} (Y;Z) - I_{e_{maj}} (Y;X) 
    \geq 
    I_{tr} (Y;Z) - I_{tr} (Y;X) 
\end{equation}

\textbf{Remarks}: 
By induction, if we find such a model after one EI step, there exists another such model in the next step, when the majority environment is the training dataset. This process repeats until a point of diminishing returns.

Therefore, we propose a simple repetition of reference model training on a biased subset of samples to get a more biased reference model better approximating $\Phi_{Spurious}$. We use ERM to approximate such a model in each step. The strategy can be defined in the following steps:
\begin{enumerate}[leftmargin=13pt,topsep=0pt,itemsep=0mm]
    \item Perform EI step from EIIL using the ERM reference model trained on the whole dataset.
    \item Instead of training an invariant learning algorithm directly on the obtained partition, retrain an ERM model on the \textit{majority} environment obtained from the previous step. Repeat this step for $n$ iterations.
    \item Use the finally obtained partition for downstream invariant learning methods. 
\end{enumerate}

The repetition of the EI step is based on the proposed conjecture that majority subsets obtained would show even stronger correlation between spurious and target labels and, the ERM model trained on it would iteratively approximate $\Phi_{Spurious}$.

\section{Experimental Results}

We show the effectiveness of our proposed method on various datasets under different challenging conditions and find that it recovers \textit{good} environment splits and provides boosted performance in invariant learning. We also provide empirical evidence to support our conjecture.

\subsection{Datasets}
\label{sec:datasets}
\textbf{Colored MNIST (CMNIST)} was originally introduced in the IRM paper \cite{irm}. It has a synthetic binary classification task where color is introduced as an anti-causal spurious correlation.

\textbf{CIFAR-Background MNIST (CBMNIST)} We introduce this new dataset, which is based on the simple concept of putting MNIST \cite{mnist} digits on CIFAR \cite{cifar} backgrounds which would act as spurious correlations. The classification task for the resultant images remains same as the MNIST digit classification task. 

\textbf{CBMNIST [I]} was created in the following manner:
\begin{itemize}[leftmargin=13pt,topsep=0pt,itemsep=0mm]
    \item One random CIFAR image from each class is chosen to act as the background for MNIST images. This results in 10 different backgrounds with different visual properties. 
    \item Each MNIST class is mapped with a randomly selected CIFAR class. This mapping between classes acts as the anti-causal spurious correlation in the training set. For the test set, a completely different random mapping of classes is chosen such that there is no overlap of digit-background combinations with the training set.
    \item Training data is split into two environments with 25000 samples each (from 60000 MNIST training samples). Test data has the remaining 10000 samples.
    \item All training samples are filled with the mapped CIFAR backgrounds. In environment 1, backgrounds of $\sim$1\% of samples are \textit{shuffled}. \textit{Shuffled} here means a random background which is not equal to the one in the train or test mapping of that class (one of the remaining 8), is applied to this sample. Similarly, in environment 2, $\sim$2\% samples have their background \textit{shuffled}.
\end{itemize}

The resulting dataset has $\sim$98.5\% samples exhibiting a strong anti-causal relationship between the target label and the background. The rest $\sim$1.5\% samples have \textit{shuffled} backgrounds so the only \textit{invariant} features in them are the digit shapes. This supports our dataset assumption (\ref{ass:ass1}) strongly, which is required for our conjecture (\ref{con:con1}) to hold.

\textbf{CBMNIST [II]} has a slight variation of the concept of \textit{shuffled}. Here \textit{shuffled} samples have their background switched randomly to exactly one other pre-decided background rather than 8 in CBMNIST [I]. In this way, even the $\sim$1.5\% \textit{shuffled} samples exhibit a spurious correlation with the background and a particular digit is only ever seen with at most 2 different backgrounds in the training set. The test set remains the same.

\textbf{CBMNIST [III]} has the percentage of samples having \textit{shuffled} backgrounds change from CBMNIST [I]. We changed the numbers 1\% and 2\% to 10\% and 20\% respectively, simulating a case where even simple ERM can focus on invariant causal features because of lots of background variations. It is unclear whether our assumption (\ref{ass:ass1}) holds in this case.


\begin{figure}[h]
	\centering
	\includegraphics[width=0.7\linewidth]{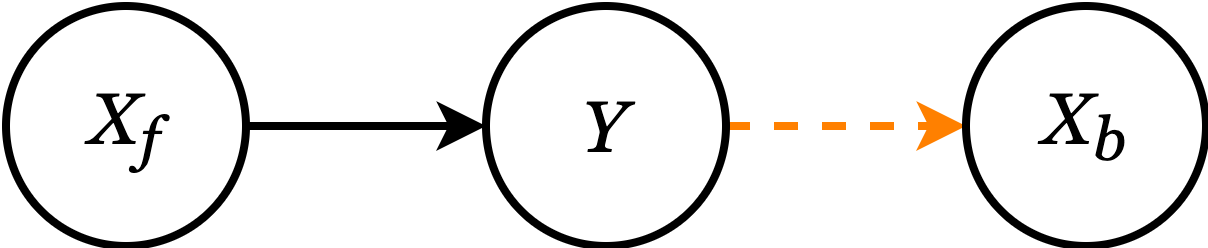}
	\caption{CBMNIST Data generating process. Here $X_f$ is the foreground or digit shape (from MNIST) that directly causes the class label $Y$ of a sample. $X_b$ is the spuriously correlated background (from CIFAR) which is caused by the label $Y$ but the mechanism generating this distribution $P(X_b|Y)$ is unstable and is represented with a colored dashed edge inspired from \cite{subbaswamy2022unifying}. Note that these variables are abstract and need to be extracted from images.}
	\label{fig:cbmnist_graph}
\end{figure}




\subsection{Implementation details}
We reuse the code [\href{https://github.com/aamixsh/reiil}{github link}], model architectures and hyperparameters from EIIL, which follows IRM for CMNIST experiments. For CMNIST, we increase the number of training steps to 900. For CBMNIST, we use an MLP with a similar structure but instead of down sampling as in CMNIST, we use all the $32 \times 32 \times 3$ input dimensions to help the model learn any complex background features. For REIIL, we found that in $n=9$ iterations, the performance usually saturates. Note that for EIIL and REIIL, the environment labels are not used in the training. We also used L2 regularization on the parameters for all methods.

\begin{figure}[bth]
	\centering
	\includegraphics[width=1\linewidth]{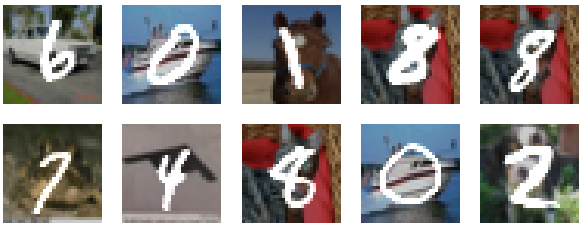}

	\caption{CBMNIST [I] samples. We use CIFAR images as the background for MNIST digits. Figure \ref{fig:cbmnist_graph} shows the data generating procss.}
	\label{fig:cbmnist_samples}
\end{figure}
\subsection{Result Analysis}

\subsubsection{Better Invariant Learning}
\begin{figure}[h]
	\centering
	\setlength{\tabcolsep}{1pt}
	\begin{tabular}{cc}
	    	\includegraphics[width=0.5\linewidth]{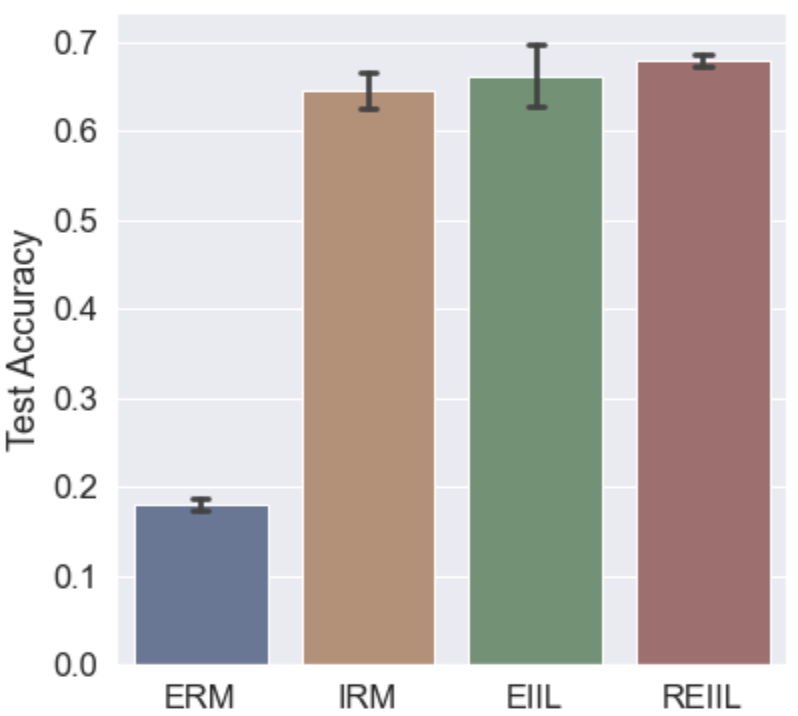} & 	\includegraphics[width=0.5\linewidth]{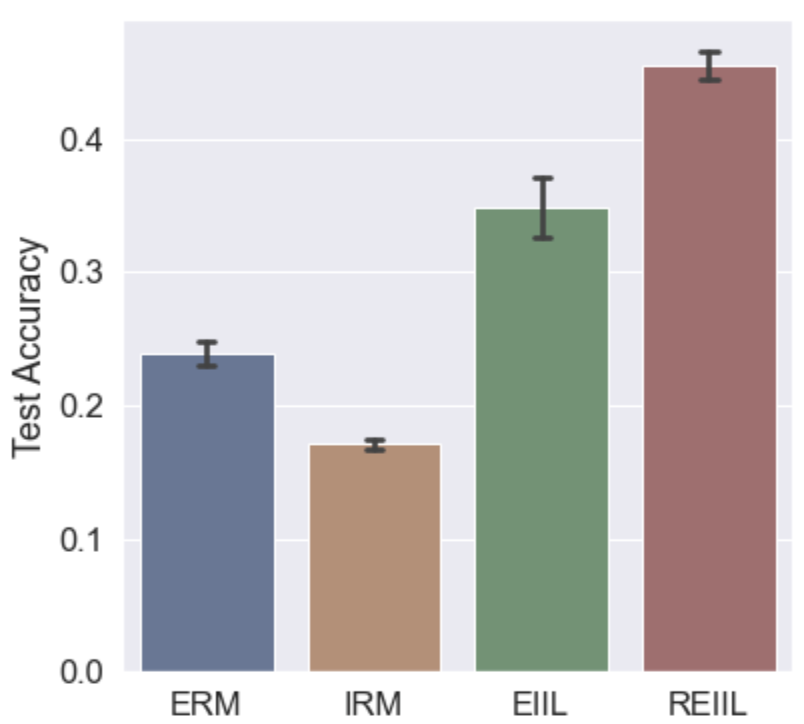}\\
	    	(a) CMNIST & (b) CBMNIST [I]\\
	    	\includegraphics[width=0.5\linewidth]{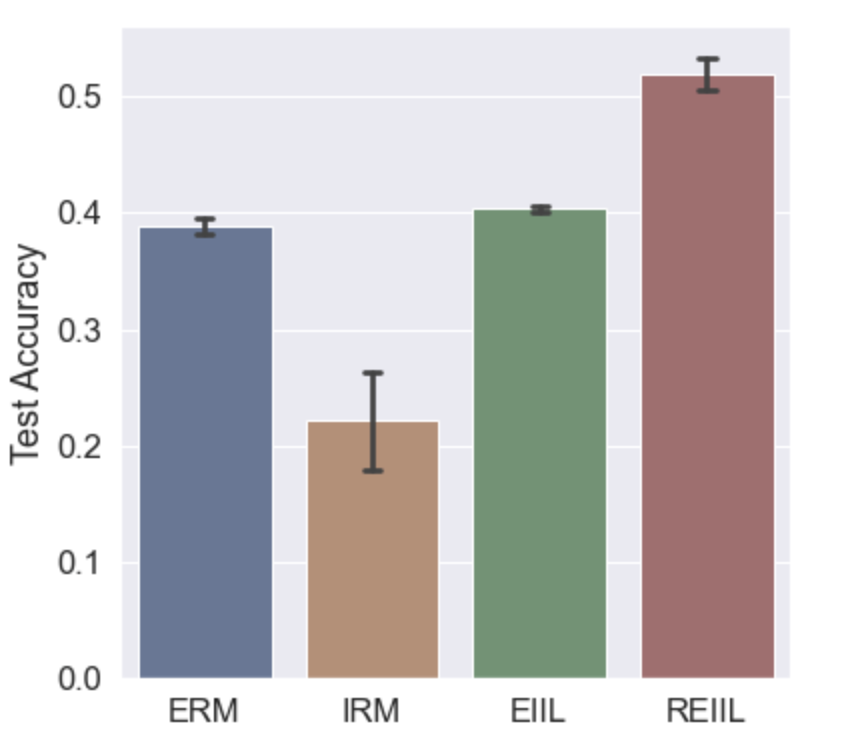} & 	\includegraphics[width=0.5\linewidth]{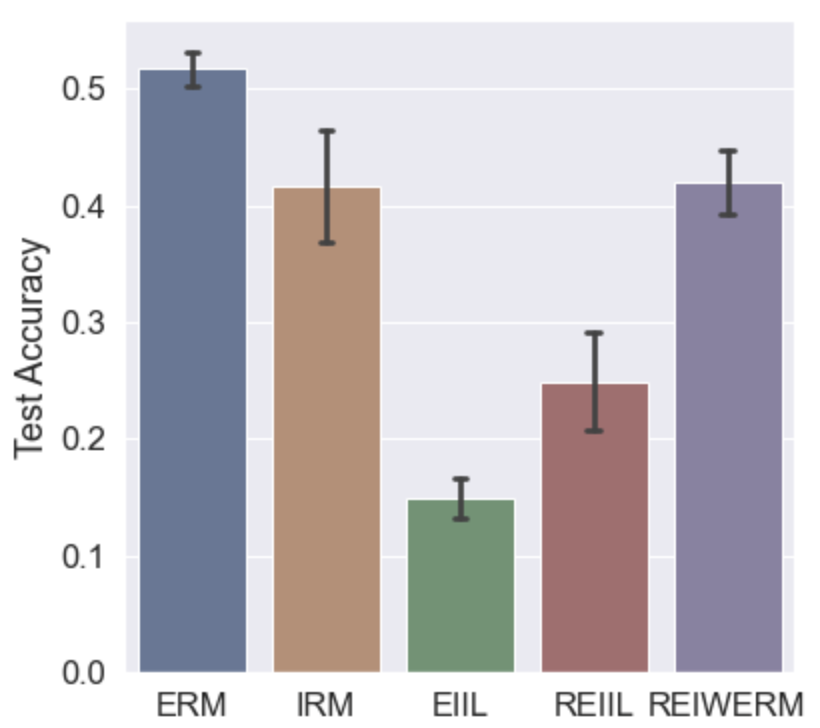} \\
	     (c) CBMNIST [II] & (d) CBMNIST [III]
	\end{tabular}
		\vspace{-10pt}
	\caption{Comparison with baseline methods on CMNIST and CBMNIST.}
	\label{fig:cmnist&cbmnist}
\end{figure}

Test performance on CMNIST and CBMNIST variants can be seen in Figure \ref{fig:cmnist&cbmnist}. In both datasets, REIIL improves over EIIL and outperforms others in most cases.

In CMNIST, the performance gains are insignificant because $\Phi_{ERM}$ approximates $\Phi_{Spurious}$ quite well in the first EIIL step itself. However, REIIL's performance is more stable than EIIL's across five runs. 

In CBMNIST [I], the performance gains are quite substantial. Because the \textit{shuffled} samples are scarce and poorly split, IRM fails to beat even ERM. But REIIL consistently performs better than other methods. For CBMNIST [II], the results were quite similar. Interestingly, in both these cases, ERM outperforms the native IRM, and using the EI step proves critical for performance gains.

ERM emerges as the best-performing model in CBMNIST [III], probably because the higher percentage of \textit{shuffled} background samples can regularize the ERM model's training. The initial reference model is also very far from $\Phi_{Spurious}$, making the performance of EIIL and even REIIL worse than IRM. REIIL still beats EIIL, but it cannot reach the performance of even IRM due to the poor starting point. It highlights the importance of a good reference model for the success of Environment Inference (\ref{eq:ei}).

One thing to note is the superior performance of the reweighting method to invariant Learning methods. Inspired from \cite{jtt}, we trained a Weighted ERM (WERM) model on the EI obtained environment splits (REIWERM). WERM weighs samples within each environment equally rather than weighing each sample equally in the whole dataset as in ERM. This model surprisingly matched the performance of IRM. We defer investigating this further for future work.

\subsubsection{Dynamics of Repeated EI} If datasets follow our assumption (\ref{ass:ass1}) and we find a model satisfying Conjecture (\ref{con:con1}) using ERM, we claim that repeating the EI step produces better environment splits such that the spurious features are more informative about the labels in the majority environment (for example, corresponding to the \textit{non-shuffled} images in CBMNIST [I]). This would imply that the minority environment's test accuracy would be low under the reference model trained using the majority environment. It would also imply that most of the \textit{causally informative} samples (in the case of CBMNIST, these would be \textit{shuffled} samples) would fall in the minority environment.

\begin{figure}[h]
	\centering
	\setlength{\tabcolsep}{1pt}
	\begin{tabular}{cc}	    	\includegraphics[width=0.5\linewidth]{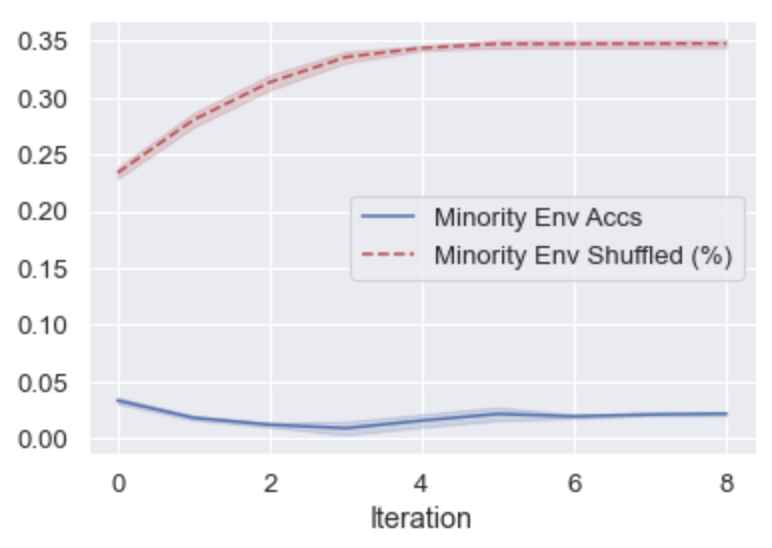} & 	\includegraphics[width=0.5\linewidth]{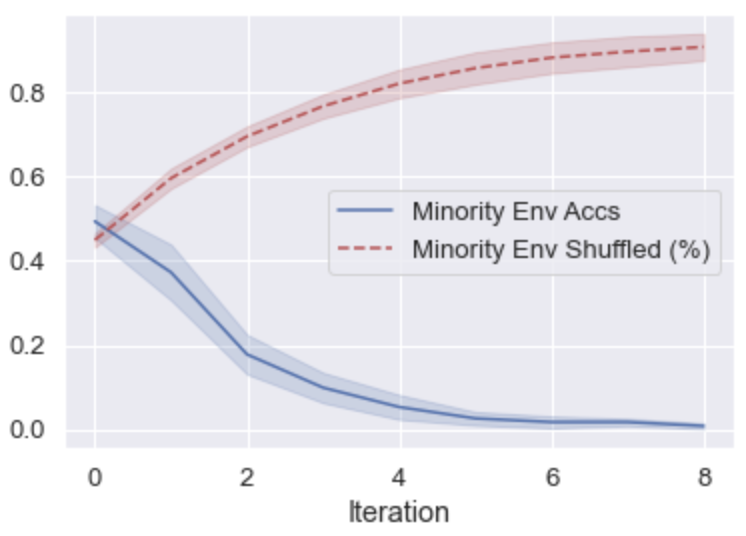} \\
	    (a) CMNIST & (b) CBMNIST [I]
	\end{tabular}
	\caption{Accuracy of reference model drops for the minority environment and percentage of \textit{shuffled} samples increase in the minority environment with REIIL iterations.}
	\label{fig:min_acc_flip}
\end{figure}

Figure \ref{fig:min_acc_flip} shows the accuracies of the reference models in each step for the samples in the \textit{minority} environment and the percentage of \textit{shuffled} samples assigned in the minority environment with each EI step. We see that accuracies keep decreasing while the proportion of \textit{shuffled} samples increases in the minority environment. Note that the training accuracies in the \textit{majority} environment always reach close to 1, implying that the ERM model successfully trains to convergence in each step. For CMNIST, \textit{shuffled} samples would correspond to those which do not exhibit the color-based spurious correlation. This effect is not as pronounced as in CBMNIST [I], but follows a similar pattern. 

Therefore, we can conclude that an ERM model trained on the majority environment obtained from the previous EI step better captures spurious features (background or color) and produces a better environment partition.

\section{Discussion}
\textbf{What if our assumption is violated?} It was shown by \cite{gulrajani2020search} that carefully trained ERM models to work quite well and often outperform Domain Generalization methods. We also see that ERM outperforms other methods in CBMNIST [III]. Our assumption is supposed to hold in more challenging situations where ERM usually fails. However, in practice, it may not be easy to test whether our assumption is valid, especially when the variables in the causal graph are hard to extract directly from data  as in the image data. We conducted experiments on synthetic data (Appendix \ref{app:exps}) where there is no clear \textit{optimal} data split due to the significant noise levels. We find that REIIL usually achieves the best test error. However, it may not recover the models close to the true ones. There also exists a model selection problem as the parameter tuning according to the validation data may make the model farther from the ground truth one. We defer more discussion to Appendix \ref{app:exps} and leave further exploration to future work. 

\textbf{What if there are more than two environments?} For simplicity, we only investigate environment inference for two environments in the paper. We are aware that a good environment split is hard to define with multiple types of spurious correlations. It is unclear whether a binary partition is sufficient. We defer further investigation of these more complicated scenarios in invariant learning for the future.

\section{Conclusion}
In this paper, we propose a novel strategy to find a \textit{more biased} reference model for environment inference. By repeating the EI step iteratively and training the reference model on the majority environment, our method helps find a better environment partition for downstream Invariant Learning tasks. Our assumption often holds in practice when most training samples exhibit strong spurious correlations and simple ERM methods fail to generalize. We conduct experiments on both CMNIST and our newly designed CBMNIST to demonstrate the effectiveness of our method.

\bibliography{paper}
\bibliographystyle{icml2022}

\newpage
\appendix

\onecolumn

\section{Synthetic Data Experiments}
\label{app:exps}

We use all variants of the synthetic data defined in \cite{irm} as well as the synthetic data defined in \cite{eiil} to test REIIL. The data is split into 3 environments with noises $0.2, 2$, and $5$ respectively. In the previous implementations, they trained ERM on all three environments, while training IRM/EIIL only on the first two and using the last one as a validation set to pick the best IRM regularizer. We change this setup by introducing a validation set explicitly using noise $= 3.5$ and treating the last environment with noise $= 5$ as the test set.

The IRM data experiment results can be summarized in Table \ref{tab:synirm}. In almost all experiments, IRM finds the best solutions. Interestingly, REIIL almost always finds models with the lowest validation and test errors (always lower than EIIL), but that does not translate to the corresponding solutions having lower Causal and Non-Causal Errors compared with the ground truth solutions.

We find that the validation set in both previous and current settings had high noise ($5$ and $3.5$), which is not representative of the real causal graph (which would have no noise). Therefore, we infer that selecting a model based on a lower validation error does not necessarily yield the best solutions. But in a learning setting with a fixed number of samples, finding a model with lower validation error is usually the best we can do. REIIL often achieves the lowest validation error, which is always lower than EIIL, suggesting that repeating the EI step improves performance. We found similar results with the synthetic data defined in EIIL, where REIIL usually found models with the lowest validation and test errors. \\

\textbf{Discussion} This questions the traditional methods of model and hyperparameter selection based on validation sets. In this case, the validation set was biased, and it is clear that the model which performs best for this set does not yield the causal solution. Even if the validation set was unbiased, it might not \textit{generalize} to biased sets because the solution does not account for the noise explicitly.

\begin{table}[h]
\centering
\resizebox{\columnwidth}{!}{
\begin{tabular}{@{}rllllllll@{}}
\toprule
Method/Type                        & FOU          & FOS          & FEU          & FES          & POU          & POS          & PEU           & PES           \\ \midrule
\cellcolor[HTML]{FFFFFF}ERM val    & 9.50 (0.26)  & 9.29 (0.23)  & 10.02 (0.40) & 9.34 (0.47)  & 10.89 (0.46) & 12.61 (1.28) & 12.29 (2.23)  & \textbf{11.68} (1.06)  \\
\cellcolor[HTML]{FFFFFF}ERM test   & 16.39 (0.76) & 16.28 (1.14) & 16.95 (1.13) & 16.01 (1.04) & 20.09 (0.95) & 22.42 (2.00) & 22.47 (5.43)  & \textbf{20.45} (2.06)  \\
\cellcolor[HTML]{FFFFFF}IRM val    & \textbf{5.11} (0.09)  & 7.66 (0.17)  & 14.84 (0.48) & 13.96 (0.68) & \textbf{6.50} (0.57)  & \textbf{8.74} (0.43)  & 16.16 (1.62)  & 16.26 (1.38)  \\
\cellcolor[HTML]{FFFFFF}IRM test   & \textbf{5.50} (0.19)  & 12.52 (0.81) & 27.12 (1.52) & 26.18 (2.09) & \textbf{9.06} (1.06)  & \textbf{13.67} (1.33) & 30.23 (3.69)  & 30.22 (2.75)  \\
\cellcolor[HTML]{FFFFFF}EIIL val   & 10.38 (0.26) & 7.57 (0.35)  & 14.01 (3.30) & 18.32 (1.39) & 10.07 (0.98) & 12.38 (2.58) & 19.01 (5.68)  & 18.78 (5.10)  \\
\cellcolor[HTML]{FFFFFF}EIIL test  & 17.67 (0.64) & 9.51 (1.51)  & 26.06 (8.6)  & 32.75 (3.43) & 18.06 (2.50) & 19.89 (4.91) & 35.37 (11.46) & 35.92 (11.33) \\
\cellcolor[HTML]{FFFFFF}REIIL val  & 7.17 (0.45)  & \textbf{6.58} (0.88)  & \textbf{8.97} (0.51)  & \textbf{8.96} (0.33)  & 8.75 (0.49)  & 10.40 (1.91) & \textbf{11.81} (2.51)  & 11.72 (1.42)  \\
\cellcolor[HTML]{FFFFFF}REIIL test & 10.77 (0.72) & \textbf{8.31} (1.96)  & \textbf{14.72} (1.29) & \textbf{15.63} (1.36) & 14.06 (1.50) & 17.09 (4.13) & \textbf{21.41} (6.00)  & 20.86 (2.97)  \\
\cellcolor[HTML]{FFFFFF}           &              &              &              &              &              &              &               &               \\
\cellcolor[HTML]{FFFFFF}ERM CE     & 0.11 (0.01)  & 0.12 (0.00)  & 0.44 (0.02)  & 0.45 (0.00)  & 0.12 (0.02)  & 0.14 (0.00)  & 0.44 (0.01)   & 0.47 (0.01)   \\
\cellcolor[HTML]{FFFFFF}ERM NCE    & 0.11 (0.00)  & -        & 0.44 (0.01)  & -        & 0.11 (0.01)  & -        & 0.42 (0.02)   & -         \\
\cellcolor[HTML]{FFFFFF}IRM CE     & \textbf{0.01} (0.00)  & 0.09 (0.01)  & \textbf{0.25} (0.01)  & \textbf{0.29} (0.01)  & \textbf{0.01} (0.00)  & \textbf{0.04} (0.01)  & \textbf{0.22} (0.02)   & \textbf{0.29} (0.01)   \\
\cellcolor[HTML]{FFFFFF}IRM NCE    & \textbf{0.01} (0.00)  & -        & \textbf{0.32} (0.00)  & -        & \textbf{0.01} (0.00)  & -        & \textbf{0.33} (0.01)   & -         \\
\cellcolor[HTML]{FFFFFF}EIIL CE    & 0.12 (0.02)  & \textbf{0.02} (0.00)  & 0.40 (0.05)  & 0.46 (0.02)  & 0.10 (0.03)  & 0.09 (0.03)  & 0.31 (0.10)   & 0.42 (0.05)   \\
\cellcolor[HTML]{FFFFFF}EIIL NCE   & 0.10 (0.02)  & -        & 0.45 (0.01)  & -        & 0.05 (0.04)  & -        & 0.39 (0.08)   & -         \\
\cellcolor[HTML]{FFFFFF}REIIL CE   & 0.05 (0.00)  & \textbf{0.02} (0.00)  & 0.49 (0.04)  & 0.45 (0.02)  & 0.05 (0.04)  & 0.09 (0.02)  & 0.42 (0.06)   & 0.47 (0.02)   \\
\cellcolor[HTML]{FFFFFF}REIIL NCE  & 0.04 (0.01)  & -        & 0.51 (0.02)  & -        & 0.04 (0.03)  & -        & 0.47 (0.05)   & -         \\ \bottomrule
\end{tabular}
}
\caption{Results of synthetic data experiments. val and test denote validation and test set errors of the model after training. CE and NCE denote Causal and Non-Causal errors. (standard deviation across 3 runs in brackets)}
\label{tab:synirm}
\end{table}


\end{document}